\documentclass[3p,10pt,number]{elsarticle}
\usepackage{amssymb}
\usepackage{amsmath}
\usepackage{xurl}
\usepackage{hyperref}
\hypersetup{
	colorlinks=true,
	breaklinks=true,
	urlcolor=blue,
	citecolor=blue,
	linkcolor=blue,
	bookmarksnumbered=true,
	bookmarksdepth=3,
	unicode=true,
	pdftitle={Institutional Equity Holdings Prediction Using Node Affinities of Dynamic Graphs},
	pdfauthor={Emad Izadifar, Zahed Rahmati},
	pdfsubject={Institutional equity holdings prediction using node affinities of dynamic graphs},
	pdfkeywords={Dynamic Graphs, Graph Neural Networks, Node Affinity Prediction, Dynamic Portfolio Allocation, Institutional Ownership Dynamics, Institutional Demand Modeling, SEC Form 13F},
	pdfcreator={LaTeX with elsarticle}
}
\usepackage{cleveref}
\usepackage{booktabs}
\biboptions{sort}

\journal{An Elsevier Journal}

\begin{document}

\begin{frontmatter}

%% Title, authors and addresses

%% use the tnoteref command within \title for footnotes;
%% use the tnotetext command for theassociated footnote;
%% use the fnref command within \author or \affiliation for footnotes;
%% use the fntext command for theassociated footnote;
%% use the corref command within \author for corresponding author footnotes;
%% use the cortext command for theassociated footnote;
%% use the ead command for the email address,
%% and the form \ead[url] for the home page:
%% \title{Title\tnoteref{label1}}
%% \tnotetext[label1]{}
%% \author{Name\corref{cor1}\fnref{label2}}
%% \ead{email address}
%% \ead[url]{home page}
%% \fntext[label2]{}
%% \cortext[cor1]{}
%% \affiliation{organization={},
%%             addressline={},
%%             city={},
%%             postcode={},
%%             state={},
%%             country={}}
%% \fntext[label3]{}

\title{Institutional Equity Holdings Prediction Using Node Affinities of Dynamic Graphs}

%% use optional labels to link authors explicitly to addresses:
%% \author[label1,label2]{}
%% \affiliation[label1]{organization={},
%%             addressline={},
%%             city={},
%%             postcode={},
%%             state={},
%%             country={}}
%%
%% \affiliation[label2]{organization={},
%%             addressline={},
%%             city={},
%%             postcode={},
%%             state={},
%%             country={}}

\author[label1]{Emad Izadifar} %% Author name
\ead{emad.izadifar@aut.ac.ir}

\author[label1]{Zahed Rahmati\corref{cor}}
\ead{zrahmati@aut.ac.ir}
\address[label1]{Department of Mathematics and Computer Science, Amirkabir University of Technology, Tehran, Iran}
\cortext[cor]{Corresponding author}

%% Abstract
\begin{abstract}
\label{abst}
%% Text of abstract
Institutional equity holdings disclosed in SEC Form 13F filings provide a rich temporal record of portfolio decisions by large investment managers. However, forecasting future allocations and modeling future demand remains challenging due to disclosure lags, reporting noise, and strong persistence in institutional behavior. We introduce the first benchmark for these tasks using temporal graph machine learning, framing holdings prediction as node affinity prediction---i.e., forecasting portfolio weights---on a discrete-time temporal bipartite graph of managers and securities extracted from preprocessed filings. On a sampled dataset comprising 99 managers and the S\&P 500 index (503 securities, 209,351 temporal edges across 48 quarters from 2013--2025), Node Affinity prediction model using Virtual State (NAVIS) achieves a state-of-the-art test Normalized Discounted Cumulative Gain (NDCG) of 0.9127 with features (0.9121 without), outperforming all dynamic graph representation learning competitors by a substantial margin, and outperforming all heuristic methods. Remarkably, a simple Exponential Moving Average baseline achieves 0.8882, surpassing all dynamic graph models except NAVIS and all heuristics except Persistent Forecast (0.8891), highlighting the strong smoothness and persistence of institutional portfolios. Domain-specific node features provide only marginal gains (<1.2\%), indicating that temporal and structural signals in the 13F ownership graph already capture most of the predictable information. By benchmarking a suite of Temporal Graph Benchmark (TGB) models under the node affinity prediction setting, both with and without features, on real-world 13F data, this work provides a reproducible foundation for temporal graph machine learning in holdings prediction and portfolio allocation. The source code is available at: \url{https://github.com/e-izdfr/portfolio-holdings-prediction}.
\end{abstract}

%%Graphical abstract
%\begin{graphicalabstract}
%	\centering
%	\includegraphics[width=\linewidth]{graphical_abstract.png}
%\end{graphicalabstract}

%%Research highlights
%\begin{highlights}
%\label{high}
%\item First application of node affinity prediction task to predict equity holdings.
%\item NAVIS dominates all dynamic graph representation learning and heuristic models.
%\item Domain-specific node features yield only slight improvements (<1.2\%).
%\item Top affinity predictions strongly predict future position changes.
%\item Establishes the first temporal graph learning benchmark for holdings prediction.
%\end{highlights}

%% Keywords
\begin{keyword}
\label{key}
%% keywords here, in the form: keyword \sep keyword
Dynamic Graphs \sep Graph Neural Networks \sep Node Affinity Prediction \sep Dynamic Portfolio Allocation \sep Institutional Ownership Dynamics \sep Institutional Demand Modeling \sep SEC Form 13F
%% PACS codes here, in the form: \PACS code \sep code

%% MSC codes here, in the form: \MSC code \sep code
%% or \MSC[2008] code \sep code (2000 is the default)

\end{keyword}

\end{frontmatter}

%% Add \usepackage{lineno} before \begin{document} and uncomment 
%% following line to enable line numbers
%% \linenumbers

%% main text
%%

%% Use \section commands to start a section
\section{Introduction}
\label{intro}
Institutional investors exercising investment discretion over more than \$100 million in Section 13(f) securities are required by the U.S. Securities and Exchange Commission to file Form 13F within 45 days of each calendar quarter-end. These filings collectively constitute one of the most comprehensive, publicly available snapshots of the U.S. equity ownership landscape \cite{sec2024form13f_readme,sec2026form13f_datasets,secform13f_reports,sec2026form13f_questions,gompers2001institutional,backus2021common}. Since their inception under the Securities Exchange Act of 1934, 13F disclosures have served as a critical resource for researchers, market participants, and regulators~\cite{boermans2026literature} seeking to understand concentration \cite{fichtner2017hidden,backus2021common}, common ownership \cite{azar2018anticompetitive,backus2021common}, and the investment behavior of large asset managers \cite{sias2006changes,gompers2001institutional}.

The rapid growth of passive indexing \cite{fichtner2017hidden,wurgler2010economic,ben2017exchange}, the concentration of assets among a handful of giant managers \cite{fichtner2017hidden,backus2021common}, and the increasing availability of large-scale longitudinal holdings data \cite{gompers2001institutional,sias2006changes} have increased the importance of Form 13F disclosures beyond their original role as a relatively static regulatory artifact \cite{sec2024form13f_readme,sec2026form13f_questions}, enabling their use as a rich temporal panel for analyzing institutional portfolio evolution \cite{gompers2001institutional,sias2006changes}. Each filing reveals not only what a manager owns at a point in time, but---when viewed sequentially---how portfolios evolve \cite{gompers2001institutional,sias2006changes}, which stocks enter or exit favor \cite{sias2006changes,yan2009institutional}, and how investment styles shift in response to market conditions \cite{wermers2012matter}. This dynamic dimension has sparked a growing literature on institutional herding \cite{lakonishok1992impact}, style drift \cite{wermers2012matter}, and the predictive power of institutional ownership changes for future returns \cite{yan2009institutional,sias2006changes}.

A particularly intriguing question is whether the historical ownership patterns embedded in successive 13F filings may contain predictive information about future institutional portfolio evolution and holdings dynamics \cite{gompers2001institutional,sias2006changes,yan2009institutional}. In other words, can we infer a latent "affinity" that each manager exhibits toward individual securities, an affinity that persists through time and manifests in the probability of initiating, increasing, or maintaining a position in the next quarter? Accurately forecasting these affinities is especially powerful because institutional investors now dominate U.S. equity markets, with almost 80\% of publicly traded equity held by institutions~\cite{barardehi2024you}, institutional ownership accounting for the majority of the market capitalization of S\&P 500 firms~\cite{lewellen2022institutional}, and ownership increasingly concentrated among a small number of giant asset managers \cite{ben2021granular,fichtner2017hidden}. Even modest improvements in predicting where this concentrated capital may flow next can translate into substantial economic signals, as changes in institutional ownership and trading behavior have been shown to generate predictable return responses \cite{yan2009institutional}. The presence of persistent behavioral biases among institutional investors further suggests that portfolio allocation decisions may exhibit systematic patterns that persist over time \cite{aren2016behavioral}. Institutional demand imbalances can create price-pressure and liquidity effects across securities \cite{koch2016commonality}. Institutional attention and information-processing frictions further contribute to predictable trading and underreaction patterns \cite{ben2017depends}. Moreover, stocks attracting large passive or index-tracking investors may experience substantial demand shocks and ownership changes \cite{appel2016passive,fichtner2017hidden}. Early detection of such shifts may therefore enable earlier identification of crowding and fire-sale risks \cite{greenwood2015vulnerable,coval2007asset}, more precise imputation of latent institutional demand in asset-pricing models \cite{koijen2019demand}, and more timely regulatory monitoring of systemic ownership concentration \cite{backus2021common,fichtner2017hidden}. If such affinities can be learned accurately, they may offer a principled way to model institutional demand, forecast ownership concentration, and disentangle genuine conviction from mechanical or index-driven trading \cite{wurgler2010economic,appel2016passive}.

Recent advances in temporal graph machine learning provide a natural framework for modeling such dynamic interaction structures \cite{feldman2026revisting,tjandra2024enhancing,ding2025dygmamba,lu2024improving,zhang2025efficient,luo2022neighborhoodaware,sarigun2023graph,poursafaei2022towards,huang2023temporal,yu2023towards,rossi2020temporal,melucci2019stock}. Managers and securities can naturally be represented as the two partitions of a bipartite graph, with possibly time-stamped ownership edges weighted by market value \cite{chen2026fund,lavin2019modeling,delpini2019systemic,melucci2019stock}. Such representations enable models to capture (i) the evolving preferences of each manager, (ii) the changing attractiveness of each security, and (iii) the complex, time-varying interactions between the two sides.

In temporal graph machine learning, dynamic link property prediction~\cite{huang2023temporal} has received the most attention as a task formulation, whereas dynamic node property prediction~\cite{huang2023temporal} has been comparatively less explored. Motivated by this imbalance, in this article, we are---to the best of our knowledge--- the first to formulate the prediction of institutional equity holdings from SEC Form 13F filings as a dynamic node property prediction task---specifically, the (dynamic) node affinity prediction subtask introduced in the Temporal Graph Benchmark~\cite{huang2023temporal}---and to benchmark it on the TGB leaderboard models \cite{tgb2026leaderboard}. We cast the task as predicting, for each institutional manager (CIK) at quarter $t$, an affinity vector over the entire universe of securities (CUSIPs) that reflects the expected market value of holdings in quarter $t+1$.

We construct a discrete-time temporal (or dynamic) heterogeneous directed, weighted bipartite graph (DTDG)~\cite{dizaji2026tgrab,shamsi2025mint,chmura2025tgm,gastinger2024tgb,huang2025utg,kazemi2020representation} from a cleaned panel of quarterly Form 13F filings spanning 2013--2025 and covering the full S\&P Global 500 securities. Manager and security nodes are augmented with rich time-varying features including turnover, sector exposures, historical similarity, institutional ownership breadth, volatility, and top holdings entropy. Following the official TGB node property prediction protocol, we train all competing models using the benchmark’s standardized data loading, temporal splits, and evaluation pipeline. To clearly isolate the impact of node features, every model is trained in two configurations: once with the full set of financial features and once in a completely featureless setting that relies solely on graph structure and temporal information.

Our experiments reveal that NAVIS~\cite{feldman2026revisting} and heuristic methods~\cite{huang2023temporal,izadifar2025ema} consistently outperform all learned temporal graph models in forecasting next-quarter holdings, as measured by Normalized Discounted Cumulative Gain at 10 (NDCG@10) \cite{huang2023temporal}. This underscores the high persistence and smoothness inherent in institutional portfolio changes, where recent allocations are highly predictive of future ones. Adding node features improves performance only slightly across all models (typically by less than 1.2\%), indicating that the temporal and structural signals in the 13F ownership graph already contain most of the predictive information for forecasting future holdings. The top-performing affinity predictions---particularly those from NAVIS and the heuristic methods---nevertheless capture persistent and economically intuitive allocation patterns, with many managers exhibiting stable preferences toward particular subsets of securities that evolve gradually over time and remain informative about future portfolio adjustments.

By framing institutional holdings prediction as a node affinity prediction problem and evaluating a suite of temporal graph models on the TGB node property prediction track---both with and without node features---this work establishes a rigorous, reproducible benchmark for applying temporal graph machine learning to institutional holdings dynamics, while providing a foundation for future research on ownership concentration, portfolio evolution, institutional herding, crowding risks, price pressure, and latent institutional demand in financial markets.

\section{Problem Definition}
\label{prob}
We study the problem of predicting future quarterly equity holdings of institutional investment managers as disclosed in SEC Form 13F filings. The proposed formulation is not restricted to Form 13F data and can be naturally applied to other institutional portfolio disclosure and regulatory filings, such as Form 13G, Form 13D, and Form N-PORT, as well as to portfolios containing both equity and non-equity asset classes.

We model institutional holdings as a discrete-time temporal heterogeneous directed, weighted bipartite graph. Let $\mathcal{M}_{t}$ and $\mathcal{S}_{t}$ denote the sets of institutional managers and securities, respectively, at quarter $t$. We define the global sets of managers and securities as
\[
\mathcal{M} = \bigcup_{t=1}^{\infty} \mathcal{M}_{t}, 
\qquad
\mathcal{S} = \bigcup_{t=1}^{\infty} \mathcal{S}_{t}.
\]
The node set at quarter $t$ and the global node set are defined as
\[
\mathcal{V}_{t} = \mathcal{M}_{t} \cup \mathcal{S}_{t},
\qquad
\mathcal{V} = \bigcup_{t=1}^{\infty} \mathcal{V}_{t}.
\]
We associate each node $u \in \mathcal{V}_{t}$ with a time-varying feature vector
\[
\mathbf{x}_{u}^{t} \in \mathbb{R}^{d_u},
\]
where $d_u$ depends only on node type (i.e., $d_u = d_{\mathcal{M}}$ if $u \in \mathcal{M}$ and $d_u = d_{\mathcal{S}}$ if $u \in \mathcal{S}$). The feature matrix at time $t$ is denoted by
\[
\mathbf{X}_{t} = \{\mathbf{x}_{u}^{t}\}_{u \in \mathcal{V}_{t}}.
\]
The discrete-time dynamic graph evolves over calendar quarters $t = 1,2,\ldots$. At each quarter $t$, the graph snapshot is
\[
\mathcal{G}_{t} = (\mathcal{V}_{t}, \mathcal{E}_{t}, \mathbf{X}_{t}),
\]
where $\mathcal{E}_{t}$ denotes the set of directed weighted manager--security edges observed at quarter $t$. An edge
\[
e_{m,s}^{t} \in \mathcal{E}_{t}
\]
connects manager $m \in \mathcal{M}$ to security $s \in \mathcal{S}$ if manager $m$ reports a position in security $s$ at quarter $t$. Each edge is associated with a non-negative weight
\[
v_{m,s}^{t} \in \mathbb{R}_{\ge 0},
\]
representing the reported market value of the position in thousands of U.S. dollars. If manager $m$ reports no position in security $s$ at quarter $t$, we define
\[
v_{m,s}^{t} = 0.
\]

In TGB, node affinity prediction is a dynamic node property prediction subtask in which the objective is to predict, for each node, a future affinity vector describing its strength of association with other nodes. For a dynamic directed weighted graph, the affinity of a node $u$ toward a node $v$ at time $t$ can be defined as the normalized outgoing edge weight
\[
a_{u,v}^{t}
=
\frac{w_{u,v}^{t}}
{\displaystyle\sum_{z \in \mathcal{V}_{t}} w_{u,z}^{t}},
\]
where $w_{u,v}^{t}$ denotes the weight of the edge from $u$ to $v$ at time $t$. If there is no directed edge going out of node $u$ into any other node at time $t$, we define
\[
a_{u,v}^{t} = 0.
\]
The affinity vector of node $u$ is then
\[
\mathbf{a}_u^{t}
=
\bigl(a_{u,v}^{t}\bigr)_{v \in \mathcal{V}_{t}},
\]
which describes how the total outgoing edge weight of $u$ is distributed among its neighbors.
In our manager--security dynamic graph, the edge weight
\[
v_{m,s}^{t}
\]
represents the reported market value of security $s$ held by manager $m$ at quarter $t$. The total outgoing edge weight of manager $m$ is therefore
\[
V_m^{t}
=
\sum_{s\in\mathcal{S}_{t}} v_{m,s}^{t},
\]
which corresponds to the manager's total reported Form 13F equity assets under management (AUM). Consequently, the affinity of manager $m$ toward security $s$ is
\[
a_{m,s}^{t}
=
\frac{v_{m,s}^{t}}
{V_m^{t}},
\]
whenever $V_m^{(t)}>0$, and $0$ otherwise.

The resulting affinity vector
\[
\mathbf{a}_m^{t}
=
\bigl(a_{m,s}^{t}\bigr)_{s\in\mathcal{S}_{t+}}
\]
is exactly the portfolio-weight vector of manager $m$, since each component represents the fraction of the manager's portfolio invested in security $s$. Therefore, under our graph construction, predicting future node affinities is mathematically equivalent to predicting future institutional equity holdings (or portfolio allocations). Given the historical graph sequence
\[
\mathcal{G}_1,\mathcal{G}_2,\ldots,\mathcal{G}_t,
\]
the node affinity prediction task becomes the prediction of
\[
\mathbf{a}_m^{t+1}
=
\bigl(a_{m,s}^{t+1}\bigr)_{s\in\mathcal{S}_{t}},
\]
for every manager $m\in\mathcal{M}_{t}$, which is precisely the problem of forecasting future institutional holdings expressed as portfolio weights.

We predict percentage weights rather than absolute dollar values because the total equity AUM $V_m^{t+1}$ is not observable at prediction time, as it is only disclosed after quarter $t+1$ concludes. Consequently, forecasting absolute holdings requires forecasting both future portfolio allocations and the manager's future total equity AUM. In contrast, percentage weights are scale-invariant and capture managerial allocation decisions independent of fund flows and market valuation effects \cite{koijen2019demand,gabaix2024granular,coval2007asset,gompers2001institutional,yan2009institutional}. While absolute holdings can be computed ex post as
\[
v_{m,s}^{t+1} = a_{m,s}^{t+1} \cdot V_m^{t+1},
\]
after the corresponding filings are released, directly modeling them conflates active portfolio rebalancing with passive changes in fund size \cite{sirri1998costly,chevalier1997risk,coval2007asset}. Therefore, we focus on predicting relative affinity vectors, which provide a more stable and economically meaningful forecasting target.

This formulation captures the normalized nature of portfolio allocation decisions and reflects the relative conviction a manager assigns to each security, independent of overall portfolio size.
\section{Related Works and Challenges}
\label{rel}
\subsection{Related Works}
\label{rel:rel}
Our problem of institutional equity holdings prediction is most closely related to institutional ownership dynamics~\cite{gompers2001institutional,sias2004institutional}, institutional demand modeling~\cite{koijen2019demand,gabaix2024granular}, and dynamic portfolio allocation \cite{markowitz1952portfolio,merton1969lifetime}. While related literature has studied institutional ownership, mutual fund trading behavior~\cite{wermers1999mutual,grinblatt1995momentum}, portfolio similarity~\cite{satone2022fund2vec,chen2026fund,lavin2019modeling}, ownership networks~\cite{konstantinov2025network,pareek2012information}, herding~\cite{sias2004institutional}, and crowding~\cite{delpini2019systemic}, relatively little work has focused on directly forecasting future institutional holdings. These topics span several interconnected strands of literature, including empirical finance and quantitative modeling~\cite{koijen2019demand,gompers2001institutional, sirri1998costly,chevalier1997risk,miori2022sec,nayanar2023interpreting}, network science~\cite{konstantinov2025network,pareek2012information}, and machine learning \cite{olorunnimbe2022deep,buczynski2021review,htun2023survey,kumbure2022machine,vancsura2025navigating,chen2023applying,chen2024deep2}.

\subsubsection{Empirical Finance and Quantitative Modeling}
Foundational finance theory originates from Markowitz's mean--variance framework~\cite{markowitz1952portfolio}, followed by the Capital Asset Pricing Model (CAPM)~\cite{sharpe1964capital} and multifactor models \cite{fama1993common,carhart1997persistence}. These theories formalize portfolio choice as a risk--return optimization problem and view portfolio weights as outcomes of expected returns, risk preferences, constraints, and market frictions \cite{markowitz1952portfolio,merton1969lifetime,sharpe1964capital}. Traditional portfolio models~\cite{markowitz1952portfolio,merton1969lifetime,sharpe1964capital} provide a theoretical basis for understanding institutional portfolio construction, while factor-based models~\cite{fama1993common,carhart1997persistence} explain cross-sectional returns and institutional tilts toward particular risk factors. Later approaches incorporate Bayesian priors and equilibrium views~\cite{black1992global} as well as dynamic portfolio choice under uncertainty \cite{merton1969lifetime}. These frameworks provide theoretical foundations for understanding institutional portfolio allocations and motivate subsequent empirical studies of institutional ownership dynamics and demand formation \cite{gompers2001institutional,sias2004institutional,koijen2019demand}.

A broad overview of institutional investors’ behavior and economic roles is synthesized in survey articles such as~\cite{drobetz2024beyond}, which review global patterns in institutional ownership, activism, ESG involvement, and market stability;~\cite{qian2023institutional}, which summarizes theoretical mechanisms underlying institutional investor behavior and the associated empirical literature; and~\cite{wang2011corporate}, which surveys the relationship between institutional ownership and corporate governance. Related empirical studies, including~\cite{chung2011corporate}, examine how heterogeneous institutional shareholders influence governance quality and firm value. These surveys provide economic and behavioral motivations underlying changes in institutional portfolios~\cite{drobetz2024beyond,qian2023institutional,wang2011corporate}. Empirical studies further document how institutional trading affects prices through herding, information transmission, and trading pressure~\cite{lakonishok1992impact,nofsinger1999herding,sias2004institutional}.

Institutional holdings also help identify phenomena such as slow-moving capital~\cite{duffie2010asset}, institutional imitation and herding~\cite{sias2004institutional}, window-dressing behavior~\cite{lakonishok1991window}, and closet indexing \cite{cremers2009active}. These works analyze holdings data to infer latent investor behavior \cite{duffie2010asset,sias2004institutional,lakonishok1991window,cremers2009active}. Early empirical research examines how institutional investors allocate capital~\cite{gompers2001institutional}, rebalance portfolios~\cite{lakonishok1992impact,sias2004institutional}, and respond to market information \cite{nofsinger1999herding,sias2004institutional}. Classic studies investigate institutional herding~\cite{lakonishok1992impact,sias2004institutional}, information-based trading~\cite{nofsinger1999herding}, and the impact of institutional ownership on asset prices \cite{gompers2001institutional}. Regression-based and panel-data econometric models are widely used to identify determinants of institutional flows~\cite{campbell2004caught,barardehi2024you}, portfolio reallocation~\cite{campbell2004caught}, and risk exposure \cite{barardehi2024you}. Such analyses are grounded in standard econometric methodologies \cite{greene2018econometric}. Although most of these works do not directly forecast future institutional holdings, they provide important insights into the mechanisms driving institutional portfolio evolution~\cite{gompers2001institutional,sias2004institutional,campbell2004caught,barardehi2024you} and therefore constitute a foundation for subsequent predictive models \cite{campbell2004caught,barardehi2024you}.

Building on this empirical literature, statistical methods have long been used to model portfolio dynamics~\cite{hamilton2020time}, while econometric methods have been widely employed to analyze institutional behavior \cite{greene2018econometric,campbell2004caught,barardehi2024you}. Regression-based models, factor analysis, and shrinkage estimators~\cite{greene2018econometric}, together with Kalman filters and state-space models~\cite{hamilton2020time}, enable inference on latent exposures and style drift~\cite{campbell2004caught} as well as risk dynamics \cite{hamilton2020time}. These approaches emphasize interpretability and economic insight~\cite{angrist2009mostly} while providing transparent statistical baselines~\cite{greene2018econometric} for modeling portfolio dynamics~\cite{hamilton2020time} and institutional demand \cite{campbell2004caught,barardehi2024you}. Regulatory disclosures such as Form 13F enabled large-scale co-holding analyses, portfolio-overlap studies, and similarity-network construction, revealing latent structure in institutional portfolios \cite{gualdi2016statistically,delpini2020portfolio,chen2026fund}. These methods reveal latent structure in institutional portfolios without relying on complex machine-learning models \cite{gualdi2016statistically,campbell2004caught,barardehi2024you}. However, classical statistical models often assume linearity, stationarity, and weak dependence across entities. Such assumptions limit their ability to capture nonlinear interactions and cross-institution dependencies that arise naturally in interconnected financial systems.

\subsubsection{Network Science}
Network science introduced a relational perspective by modeling financial systems as graphs~\cite{kenett2015network,konstantinov2025network}, where nodes represent institutions or assets and edges encode relationships such as ownership~\cite{enriques2019institutional}, co-holding and cross-holding~\cite{anton2014connected,gualdi2016statistically}, buying and selling interactions~\cite{gong2020institutional}, or similarity \cite{chen2026fund}. Studies construct ownership networks~\cite{enriques2019institutional}, co-holding and portfolio-overlap networks~\cite{anton2014connected,gualdi2016statistically}, liquidity-dependency networks~\cite{kenett2015network}, and counterparty-exposure networks~\cite{nier2007network} to examine contagion pathways~\cite{battiston2012debtrank,acemoglu2015systemic}, fire-sale amplification~\cite{gualdi2016statistically}, market concentration~\cite{huseynov2025firmographica}, stock-level idiosyncratic volatility~\cite{zhai2024institutional}, and structural vulnerabilities in financial systems \cite{konstantinov2025network,nier2007network}. These works have employed both static and dynamic network analysis~\cite{miori2024network,demirel2024examination}, including centrality measures~\cite{kenett2015network}, community detection~\cite{gualdi2016statistically}, percolation analysis~\cite{konstantinov2025network}, and contagion modeling~\cite{battiston2012debtrank,acemoglu2015systemic}, to study shock propagation~\cite{battiston2012debtrank}, market interconnectedness~\cite{kenett2015network,enriques2019institutional}, financial stability \cite{acemoglu2015systemic,uddin2022network}, and corporate financial frictions such as debt maturity mismatch driven by investor network centrality \cite{liu2025institutional}. Pioneering systemic-risk frameworks~\cite{battiston2012debtrank,acemoglu2015systemic} formalize how network topology can amplify shocks throughout the financial system. Empirical analyses of co-holding networks~\cite{anton2014connected}, portfolio-overlap networks~\cite{gualdi2016statistically}, and fund similarity structures~\cite{chen2026fund} demonstrate how shared exposures contribute to correlated losses and systemic fragility \cite{anton2014connected,gualdi2016statistically}. These approaches primarily provide structural insights and focus on system-level effects rather than directly forecasting specific holdings \cite{battiston2012debtrank,acemoglu2015systemic,gualdi2016statistically}. Nevertheless, they establish that institutional portfolios are embedded in complex relational structures~\cite{anton2014connected,gualdi2016statistically,enriques2019institutional}, thereby motivating graph-based representations and network-aware predictive models for institutional holdings forecasting.

\subsubsection{Machine Learning}
Machine learning expanded predictive capabilities in finance by introducing nonlinear and data-driven models across a wide range of financial tasks~\cite{buczynski2021review,vancsura2025navigating}, including asset pricing~\cite{chen2024deep,gu2020empirical}, risk prediction~\cite{khandani2010consumer}, institutional investment behavior and holdings analysis~\cite{van2024machine,saxena2025predicting}, return forecasting~\cite{kumbure2022machine,chen2024deep2}, and portfolio evolution \cite{thomaz2024case}. Machine-learning methods such as random forests and gradient boosting~\cite{gu2020empirical,khandani2010consumer}, collaborative filtering~\cite{chung2025mean,sankar2015trust,roy2022systematic}, autoencoders~\cite{chen2024deep2,cao2025deep}, and deep neural networks~\cite{chen2024deep,gu2020empirical} have been successfully applied to financial forecasting and investment decision-making \cite{htun2023survey,thomaz2024case,chen2023applying,van2024machine,saxena2025predicting}. Sequence models, including recurrent neural networks, temporal convolutional networks, and Transformers~\cite{chen2024deep2,chen2024deep}, further capture complex temporal dependencies in market data and disclosure sequences \cite{thomaz2024case}. Reinforcement-learning approaches have also been explored for dynamic portfolio allocation and sequential investment decisions \cite{jiang2017deep,anisha2025aiml}. Several recent studies have moved closer to the institutional holdings prediction setting by leveraging SEC Form 13F disclosures for portfolio construction~\cite{chen2023applying}, investment recommendation~\cite{van2024machine}, institutional trading prediction~\cite{saxena2025predicting}, and next-portfolio prediction \cite{thomaz2024case}. Despite these advances, most machine-learning applications in finance remain focused on forecasting returns~\cite{kumbure2022machine,chen2024deep2}, prices and risk measures~\cite{khandani2010consumer,gu2020empirical}, or portfolio performance~\cite{thomaz2024case} rather than directly predicting future institutional holdings. Nevertheless, their ability to learn nonlinear relationships~\cite{gu2020empirical,chen2024deep}, capture high-dimensional dependencies~\cite{chen2024deep,chen2024deep2}, and exploit large-scale financial datasets~\cite{cao2025deep,kumbure2022machine} makes them a natural foundation for institutional holdings forecasting.

ML methods capture complex nonlinear dependencies that are often difficult to represent using classical econometric and statistical models~\cite{gu2020empirical,chen2024deep} and generally prioritize predictive accuracy. However, most conventional machine-learning approaches do not explicitly represent the relational structure among institutions and assets, instead capturing interactions indirectly through features, correlations, or shared covariates \cite{wang2021review,huang2023temporal}.
 
Static and dynamic graph machine learning extend this relational perspective on investors, assets, and exposures by modeling financial systems as static or temporal graphs \cite{wang2021review,konstantinov2025network}. Financial systems are often represented as bipartite or heterogeneous graphs, where institutions and assets are modeled as nodes \cite{yang2023common,konstantinov2025network,nath2023learning}. Relationships such as ownership~\cite{konstantinov2025network}, co-holding~\cite{yang2023common,saxena2022holder}, buying and selling interactions~\cite{melucci2019stock}, and similarity~\cite{satone2022fund2vec,cavar2018mapping,guo2024comparing} are represented as edges. Within this framework, institutional holdings forecasting can naturally be formulated as a link prediction problem~\cite{hamilton2020graph,huang2023temporal} or a relational recommendation task \cite{wang2021review}. However, most existing graph learning studies in finance formulate prediction as a link prediction problem in static or temporal graphs \cite{zhang2015link,wang2021review,huang2023temporal}. Representative applications include stock recommendation~\cite{melucci2019stock}, institutional holder prediction~\cite{saxena2022holder}, investor funding decision prediction~\cite{liang2016predicting}, financial representation learning~\cite{xiang2022temporal,jeyaraman2024temporal,nath2023learning}, portfolio and investment network forecasting~\cite{dai2025griffinnet,jiang2025mutual}, and financial knowledge graph reasoning \cite{guo2025evaluating,cho2024fishnet,yang2020identifying}. Node prediction tasks have received comparatively less attention in financial graph learning \cite{wang2021review}, particularly for forecasting future institutional portfolio allocations in institution--asset networks. Examples of node prediction applications include stock movement prediction~\cite{qian2024mdgnn,pacreau2021graph}, risk and market-state prediction~\cite{jiang2023network,kim2025multi}, financial representation learning~\cite{fu2025low}, asset price prediction~\cite{gu2024dystage,uddin2021attention,squartini2017enhanced}, and knowledge-graph-based financial forecasting \cite{uygun2025financial,abbas2024knowledge}. These studies demonstrate that financial entities are embedded in rich relational structures~\cite{konstantinov2025network,wang2021review,niu2024evaluating} and that graph-based learning can exploit such dependencies for prediction. Consequently, graph machine learning provides a natural framework for institutional holdings forecasting, where future portfolio decisions depend not only on entity-specific attributes but also on evolving relationships among institutions and assets. Static and dynamic hypergraph machine learning, which extends graph representations by modeling higher-order relationships among multiple entities, has also recently been applied to financial prediction tasks \cite{chen2025extracting}.

Overall, existing literature provides important insights into institutional portfolio behavior, financial network structure, and predictive modeling. Nevertheless, several limitations remain. First, most empirical finance studies focus on explaining institutional behavior rather than forecasting future holdings. Second, network science approaches primarily analyze structural properties and systemic effects rather than making asset-level predictions. Third, machine learning and graph learning studies in finance predominantly target stock prediction, recommendation, market forecasting, or financial link prediction tasks.

Direct prediction of future institutional equity holdings in dynamic institution--asset networks remains relatively underexplored, particularly from a node prediction perspective that seeks to forecast future portfolio allocations. This gap motivates the development of predictive models that jointly leverage temporal portfolio evolution and relational dependencies among institutions and assets. In this work, we directly forecast future institutional equity holdings in large-scale dynamic institution--asset networks.

\subsection{Challenges}
\label{rel:chal}
Forecasting institutional equity holdings faces structural, statistical, and computational challenges.

\subsubsection{Low-Frequency, Incomplete, and Delayed Reporting}
Institutional holdings data (e.g., Form 13F filings) are reported quarterly, filed with delays of up to 45 days after quarter-end, and may be incomplete. This limits temporal resolution, obscures intraperiod trading activity and short positions, and makes predictions partially backward-looking, complicating any dynamic analysis, whether statistical, econometric, network-science-based, or machine-learning-based \cite{sec2026form13f_datasets,sec2024form13f_readme,secform13f_reports,anderson2016form,anderson2018examination}. In addition, many positions are exempt from reporting or subject to regulatory lags, and reporting practices may vary across jurisdictions, creating further blind spots in observed institutional exposures \cite{anderson2016form,anderson2018examination}.

\subsubsection{Sparsity and Staleness}
Institutional holdings data are highly sparse, as only a small fraction of all possible manager--equity pairs correspond to observed positions at any reporting period. Moreover, institutional portfolios often exhibit substantial persistence across consecutive quarters, resulting in relatively infrequent changes in many holdings \cite{gompers2001institutional,sias2004institutional}. These characteristics create challenges for predictive modeling by introducing severe class imbalance and making meaningful portfolio changes difficult to distinguish from routine position maintenance.

\subsubsection{Data Quality and Identifier Drift}
Corporate actions, reporting inconsistencies, and identifier changes introduce significant data quality challenges in constructing consistent historical datasets \cite{anderson2016form,anderson2018examination}. In particular, identifier drift arising from changes in CUSIP and ticker codes, mergers, delistings, and identifier reuse complicates the alignment of holdings over time. Reported positions may also not always align perfectly with SEC-provided datasets due to processing differences and reporting inconsistencies \cite{sec2026form13f_datasets,sec2024form13f_readme,secform13f_reports}. Amended 13F filings may further introduce revisions that complicate the reconstruction of historical holdings \cite{anderson2016form,anderson2018examination}. As a result, constructing reliable time-consistent snapshots requires substantial preprocessing independent of downstream modeling.

\subsubsection{Changing Universe of Securities and Institutions}
The set of reporting institutional investors and securities in 13F filings is inherently time-varying, as new entities enter the reporting system while others exit over time \cite{sec2026form13f_datasets}. This changing universe complicates longitudinal analysis, including panel construction, cross-time comparability, and standard econometric assumptions such as balanced panels and stationarity. It may also affect descriptive statistics due to entry and exit dynamics \cite{greene2018econometric}.

\subsubsection{Survivorship Bias}
Survivorship bias arises in longitudinal analyses if delisted securities and inactive funds are excluded from the analysis \cite{angrist2009mostly}. This can lead to systematically biased inference in institutional holdings studies due to non-random sample truncation over time \cite{greene2018econometric}.

\subsubsection{Endogeneity and Causal Interpretation}
Institutional trading behavior affects asset prices, while prices in turn influence future portfolio allocation decisions, creating a two-way feedback loop that induces endogeneity and complicates causal inference. Addressing this issue typically requires instrumental variables, structural models, or quasi-experimental designs \cite{angrist2009mostly}. In the absence of strong identification strategies, many computational approaches instead focus on predictive performance rather than causal identification, relying on assumptions about temporal dependence, latent structure, or relational dynamics that may be affected by non-stationarity and regime shifts over time.

\subsubsection{Lack of Standardized Benchmarks}
To the best of our knowledge, there is no universally accepted benchmark dataset for institutional portfolio evolution and prediction. Differences in preprocessing pipelines, reporting formats, aggregation procedures, survivorship-bias handling, and filtering criteria limit reproducibility and hinder systematic comparison across statistical, econometric, network science, and machine learning approaches.

\subsubsection{Model Interpretability, Transparency, and Policy Relevance}
Interpretability remains an open challenge in modeling institutional trading behavior. Across econometric, network-based, and machine learning approaches, understanding the drivers of predicted portfolio decisions is often non-trivial despite differences in model structure and complexity. However, regulators, risk managers, and practitioners require explanations in addition to predictions, making interpretability an important requirement for real-world deployment \cite{ribeiro2016should}.

\subsubsection{Scalability and Computational Complexity}
Large ownership matrices or tensors (thousands of institutions $\times$ tens of thousands of securities $\times$ multiple time periods) introduce significant computational and memory challenges. Network-based and machine learning approaches face scalability constraints in large-scale graph settings, while econometric models struggle with high-dimensional panel structures \cite{konstantinov2025network,hamilton2017inductive}.

\subsubsection{Conceptual Gaps Between Disciplines}
Empirical finance, econometrics, network science, and machine learning often rely on different assumptions, evaluation protocols, and notions of dependence and causality. As a result, integrating insights across these domains remains challenging, particularly when models adopt incompatible views of dependence structure, causal inference, or equilibrium versus predictive objectives.

\subsubsection{Economic Validation}
While accurately predicting changes in institutional holdings is of methodological interest, a key empirical validation criterion is whether the resulting implied trades generate positive risk-adjusted returns after accounting for transaction costs, including turnover and market impact.

Overall, the forecasting of institutional equity holdings is challenged by limitations in data availability and quality, sparsity in observed institution--asset interactions, and the dynamic nature of both market participants and asset universes. These challenges are further compounded by scalability constraints in high-dimensional relational settings and the need to reconcile heterogeneous methodological perspectives across empirical finance, econometrics, network science, and machine learning. Collectively, they highlight the need for models capable of capturing temporal dynamics and rich relational structures in institutional portfolios, while supporting reliable prediction in large-scale dynamic networks.

In this work, we seek to account for several of the key challenges identified in the literature, including low-frequency and delayed reporting, sparsity and staleness in holdings data, data quality issues and identifier drift, the evolving universe of securities and institutions, survivorship bias, lack of standardized benchmarks, and scalability constraints in high-dimensional settings. Rather than treating these issues in isolation, our modeling and data construction pipeline is designed to mitigate their combined impact, enabling more reliable forecasting of institutional equity holdings in large-scale, time-evolving institution--asset networks.

\section{Methodology}
\label{meth}
As discussed in~\Cref{prob}, we formulate the prediction of future institutional equity holdings as a node affinity prediction task on a discrete-time temporal heterogeneous directed, weighted bipartite graph constructed from quarterly Form 13F filings. Both manager and security nodes are equipped with rich, time-varying features engineered from historical 13F data.

We train predictive models to estimate future node affinities of investment managers towards assets. Depending on the model class, the loss function is either cross-entropy loss or a customized loss function introduced in~\cite{feldman2026revisting}. All experiments are conducted using the official TGB codebase with our extensions including adding node features, facilitating reproducibility.

Model performance is evaluated using NDCG@10, which measures the quality of the predicted asset rankings with respect to actual future holdings. Higher NDCG@10 values indicate better alignment between predicted and observed institutional portfolio allocations.

\subsection{Temporal Bipartite Graph Construction}
\label{meth:const}
The raw Form 13F filings from 2013--2025 are first processed by joining the relevant filing tables and retaining the attributes required for graph construction and feature engineering. Standard preprocessing steps are then applied to clean and harmonize the data across reporting periods. To identify the target asset universe, we first retain common-stock holdings from the merged dataset. Fuzzy string matching is then used to match reported securities to the constituents of the S\&P 500 index. Matches with a similarity score below 80 are discarded, resulting in a set of 503 securities represented by their CUSIP identifiers. A random sample of 99 institutional investment managers (CIKs) is subsequently selected from the reporting population. For each quarter, we construct a graph snapshot from the holdings reported during that period, and the resulting discrete-time dynamic graph is obtained as the sequence of subgraphs induced by the selected managers and securities.

From the cleaned panel of Form 13F disclosures spanning 2013--2025, we construct a temporal bipartite graph $\mathcal{G} = (\mathcal{V}, \mathcal{E})$ with time $t$ snapshot given by $\mathcal{G}_{t} = (\mathcal{V}_{t}, \mathcal{E}_{t}, \mathbf{X}_{t})$ where
\begin{itemize}
  \item $\mathcal{M}_{t}$ is the set of 99 institutional investment managers (CIKs),
  \item $\mathcal{S}_{t}$ is the set of 503 S\&P 500 securities (CUSIPs),
\end{itemize}
and $\mathbf{X}_{t}$ is the 602 $\times$ 23 node-feature matrix at time $t$ discussed in~\Cref{meth:feat}.

\subsection{Node Features}
\label{meth:feat}
Both manager and security nodes are equipped with rich, time-varying feature vectors engineered from historical 13F data. The complete node-feature representation at time $t$ is denoted by $\mathbf{X}_t \in \mathbb{R}^{602 \times 23}$, where 602 corresponds to the total number of nodes (99 managers and 503 securities) and 23 denotes the full feature dimension after encoding.

Since the feature space is shared across both node types, features not applicable to a given node type are explicitly set to zero, yielding a shared feature space with structured zero-padding across heterogeneous node types.

\textbf{Manager (CIK) features} (17 dimensions):
\begin{itemize}
  \item \texttt{num\_stocks}: number of distinct positions held (log-transformed)
  \item \texttt{AUM}: total reported equity assets under management (log-transformed)
  \item \texttt{HHI}: Herfindahl--Hirschman Index measuring portfolio concentration (z-scored)
  \item One-hot encoded sector exposures (11 GICS sectors):
  \begin{itemize}
  	\item \texttt{sector\_Basic\_Materials},
  	\item \texttt{sector\_Communication\_Services},
  	\item \texttt{sector\_Consumer\_Cyclical},
  	\item \texttt{sector\_Consumer\_Defensive},
  	\item \texttt{sector\_Energy},
  	\item \texttt{sector\_Financial\_Services},
  	\item \texttt{sector\_Healthcare},
  	\item \texttt{sector\_Industrials},
  	\item \texttt{sector\_Real\_Estate},
  	\item \texttt{sector\_Technology},
  	\item \texttt{sector\_Utilities}
  \end{itemize}  
  \item \texttt{turnover\_ratio}: quarter-over-quarter portfolio turnover (z-scored)
  \item \texttt{historical\_similarity}: cosine similarity of current portfolio weights to past portfolios (z-scored)
  \item \texttt{top\_holdings\_entropy}: Shannon entropy of the top 10 holdings’ weights (z-scored)
\end{itemize}

\textbf{Security (CUSIP) features} (6 dimensions):
\begin{itemize}
  \item \texttt{num\_holders}: number of distinct reporting managers (z-scored)
  \item \texttt{total\_institutional\_ownership}: total reported market value across all managers (z-scored)
  \item \texttt{total\_shares\_held}: total number of shares held by institutions (z-scored)
  \item \texttt{avg\_holding\_per\_manager}: average position size per manager (z-scored)
  \item \texttt{volatility\_of\_holdings}: standard deviation of position sizes across managers (z-scored)
  \item \texttt{sector}: GICS sector (treated as a categorical feature and encoded via a learnable embedding)
\end{itemize}

Features are either log-transformed or standardized (zero mean, unit variance) using statistics computed from the training period, one-hot encoded, or encoded via learned embeddings.

\subsection{Task Formulation and Evaluation Protocol}
\label{meth:eval}
As discussed in~\Cref{prob}, we formulate institutional equity holdings prediction as a node affinity prediction task on the constructed temporal bipartite graph. For each manager $m$ at quarter $t$, the objective is to predict the affinity vector $\hat{\mathbf{a}}_m^{t+1} \in [0,1]^{503}, \text{ where } \sum_{i=1}^{503}\hat a_{m,i}^{t+1}=1,$ representing the predicted percentage allocation of manager $m$'s equity portfolio across all 503 S\&P 500 securities in the subsequent quarter $t+1$.

We adopt the official TGB temporal data split protocol (70\% training, 15\% validation, and 15\% test) \cite{huang2023temporal,tgb2026leaderboard}:
\begin{itemize}
  \item \textbf{Training}: April 2013 -- March 2023 (40 quarters, 70\% of the temporal edges)
  \item \textbf{Validation}: April 2023 -- March 2024 (4 quarters, 15\% of the temporal edges)
  \item \textbf{Test}: April 2024 -- March 2025 (4 quarters, 15\% of the temporal edges)
\end{itemize}

Performance is evaluated using NDCG@10, the standard metric employed in the TGB node property prediction benchmark. NDCG@10 is computed for each manager and then averaged across all prediction instances. The metric evaluates the quality of the predicted ranking of securities relative to the manager's actual future portfolio allocation. Higher NDCG@10 values indicate better agreement between predicted and observed institutional holdings.

\subsection{Models}
\label{meth:mod}
We evaluate models from the TGB node property prediction leaderboard~\cite{huang2023temporal,tgb2026leaderboard} and additionally include Exponential Moving Average (EMA) \cite{izadifar2025ema}, a statistical baseline based on exponentially weighted moving averages of past weights:
\begin{itemize}
  \item \textbf{Persistent Forecast} (PF)~\cite{huang2023temporal}
  \item \textbf{Moving Average} (MA)~\cite{huang2023temporal}
  \item \textbf{Exponential Moving Average}~\cite{izadifar2025ema}
  \item \textbf{DyRep}~\cite{trivedi2018representation}
  \item \textbf{TGN}~\cite{rossi2020temporal}
  \item \textbf{DyGFormer}~\cite{yu2023towards}
  \item \textbf{TGNv2}~\cite{tjandra2024enhancing}
  \item \textbf{NAVIS}~\cite{feldman2026revisting}
\end{itemize}

Unlike most existing temporal graph representation learning models, which were primarily designed for link prediction, NAVIS is specifically developed for the node affinity prediction task. Its architecture combines per-node memory with a learnable virtual global state that evolves jointly with the dynamic graph, enabling it to capture both local node histories and global temporal dynamics. Inspired by state-space models, NAVIS incorporates long-range temporal memory while retaining the flexibility of graph representation learning. In addition, it employs a ranking-based loss function that is better aligned with node affinity prediction than the cross-entropy loss commonly used for link prediction. These design choices have enabled NAVIS to achieve state-of-the-art performance on the TGB node affinity prediction benchmark \cite{feldman2026revisting,huang2023temporal}.

While additional node affinity prediction models~\cite{wu2026scadyg,panyshev2025never,lee2025simple,berndt2025permutation,qin2026learning,huang2026g,qian2025edge}
and alternative evaluation metrics~\cite{li2025test}
have been proposed in the literature, we focus on the official TGB models and the above additional baseline to facilitate reproducibility and direct comparability with the established benchmark.

To isolate the impact of node features---which are specifically engineered for our 13F institutional ownership data, unlike the original featureless TGB datasets---each model is trained under two configurations:
\begin{itemize}
  \item \textbf{With features}: using the full manager and security feature vectors described above
  \item \textbf{Without features}: using only structural and temporal information
\end{itemize}

\subsection{Training and Reproducibility}
\label{meth:train}
All experiments are conducted using the official TGB codebase and evaluation pipeline to facilitate reproducibility and direct comparability with published leaderboard results. Hyperparameters follow the recommended settings from the respective model repositories unless otherwise stated. The temporal data split follows the protocol described in \Cref{meth:eval}.

\section{Experimentation}
\label{exper}
We conduct a comprehensive empirical evaluation on a cleaned panel of quarterly SEC Form 13F filings spanning 2013--2025. The dataset covers the S\&P 500 constituent universe (503 distinct CUSIPs) and a random sample of 99 institutional investment managers (CIKs), yielding 209,351 time-stamped directed edges across 48 quarters. We follow the official TGB protocol for evaluation, reproducibility, and data splitting. The main results are reported in~\Cref{exper:rel}, and per-model visualizations are provided in \Cref{exper:vis}.

\subsection{Dataset}
\label{exper:data}
The dataset construction process is discussed in~\Cref{meth:const}, and its statistics are summarized in \Cref{exper:data:tab}.
\begin{table}[htbp]
	\centering
	\begin{tabular}{lr}
		\toprule
		\textbf{Statistic} & \textbf{Value} \\
		\midrule
		Number of manager nodes (CIKs) & 99 \\
		Number of security nodes (CUSIPs) & 503 \\
		Number of temporal edges & 209,351 \\
		Number of quarters & 48 \\
		Average edges per quarter & 4,361 \\
		Time span & Q2 2013 -- Q1 2025 \\
		\bottomrule
	\end{tabular}
	\caption{Dataset summary}
	\label{exper:data:tab}
\end{table}

\subsection{Experimental Setup}
\label{exper:set}
We follow the official TGB protocol~\cite{huang2023temporal,tgb2026leaderboard}:
\begin{itemize}
	\item \textbf{Training}: Q2 2013 -- Q1 2023 (40 quarters, 70\% of the temporal edges)
	\item \textbf{Validation}: Q2 2023 -- Q1 2024 (4 quarters, 15\% of the temporal edges)
	\item \textbf{Test}: Q2 2024 -- Q1 2025 (4 quarters, 15\% of the temporal edges)
\end{itemize}

This chronological split preserves temporal ordering and simulates a streaming evaluation setting~\cite{huang2023temporal}. All models are trained using the official TGB codebase~\cite{tgb2026leaderboard}. To support the proposed dataset and experimental setting, we extend both the TGB package and the corresponding model repositories with additional functionality, including support for node features, improved and consistent logging and visualization across models, both locally and on Weights \& Biases, as well as robust logging that is synchronized to the cloud in case of interrupted connections. Hyperparameters follow the recommended settings from the respective model repositories unless otherwise stated. We report NDCG@10 (higher is better) and cross-entropy loss. The cross-entropy loss is recorded per epoch and computed as the average across all batches.

\subsection{Main Results}
\label{exper:rel}
Key observations from~\Cref{exper:rel:tab}:
\begin{itemize}
	\item \textbf{NAVIS\_with\_features} achieves the highest Test NDCG@10 of 0.9127.
	\item \textbf{NAVIS} is, by a large margin of at least 21.46\%, the best dynamic graph representation learning model without node features, achieving a Test NDCG@10 of 0.9121.
	\item \textbf{EMA} achieves a Test NDCG@10 of 0.8882, outperforming by at least 19.07\% all dynamic graph representation learning models except NAVIS, as well as all heuristic methods except PF, which exceeds it by only 0.09\%. This is notable given the simplicity of the method.
	\item \textbf{PF} achieves a Test NDCG@10 of 0.8891, outperforming by at least 19.16\% all dynamic graph representation learning models except NAVIS, and outperforming all other heuristic methods. This highlights the strong persistence of institutional portfolio allocations, where recent holdings remain highly predictive of future ones.
	\item \textbf{Node features} provide only modest improvements---less than 1.2\% across all models---suggesting that temporal portfolio dynamics and graph structure already capture most of the predictive information available in the dataset.
\end{itemize}

\begin{table}[tbp]
	\centering
	\begin{tabular}{lccc}
		\toprule
		Model & Test NDCG@10 & Validation NDCG@10 & Epochs \\
		\midrule
		NAVIS\_with\_features & \textbf{0.9127} & 0.8818 & 500 \\
		NAVIS & 0.9121 & \textbf{0.8886} & 50 \\
		EMA & 0.8882 & 0.8762 & 48 \\
		PF & 0.8891 & 0.8691 & 48 \\
		MA & 0.8464 & 0.8433 & 48 \\
		TGNv2\_with\_features & 0.6975 & 0.7348 & 500 \\
		TGNv2 & 0.6962 & 0.7224 & 50 \\
		TGN\_with\_features & 0.6146 & 0.5714 & 50 \\
		TGN & 0.6178 & 0.5673 & 50 \\
		DyGFormer\_with\_features & 0.5965 & 0.5728 & 50 \\
		DyGFormer & 0.5857 & 0.5587 & 50 \\
		DyRep\_with\_features & 0.5804 & 0.5481 & 30 \\
		DyRep & 0.5800 & 0.5477 & 30 \\
		\bottomrule
	\end{tabular}
	\caption{Performance comparison of TGB leaderboard models and an additional EMA baseline on validation and test sets}
	\label{exper:rel:tab}
\end{table}

\subsection{Per-Model Detailed Visualizations}
\label{exper:vis}
For transparency, we provide NDCG@10 performance visualizations for each model over time.
% NAVIS_with_features
\begin{figure}[tbp]
	\centering
	\includegraphics[width=\linewidth]{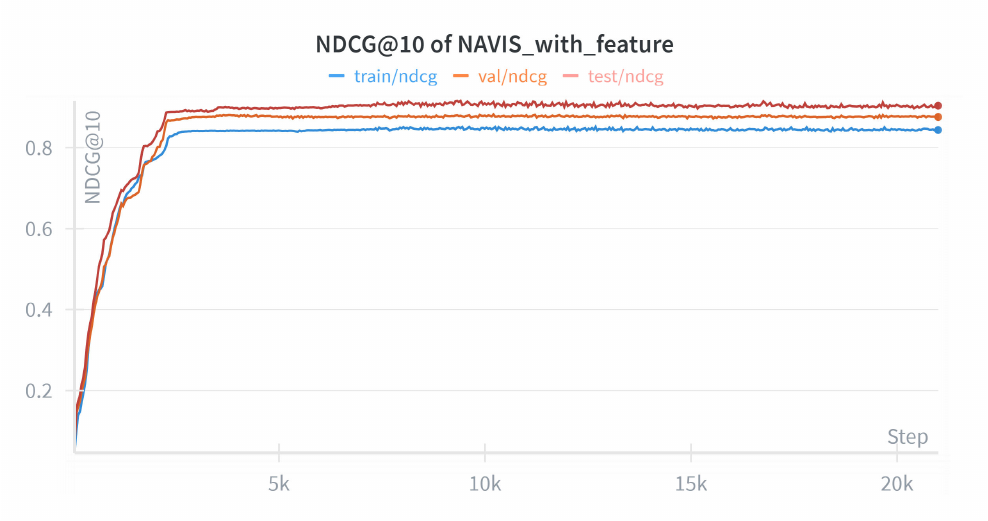}
	\caption{NDCG@10 over time for NAVIS with node features}
	\label{exper:vis:navisfeat}
\end{figure}

% Exponential Moving Average
\begin{figure}[tbp]
	\centering
	\includegraphics[width=\linewidth]{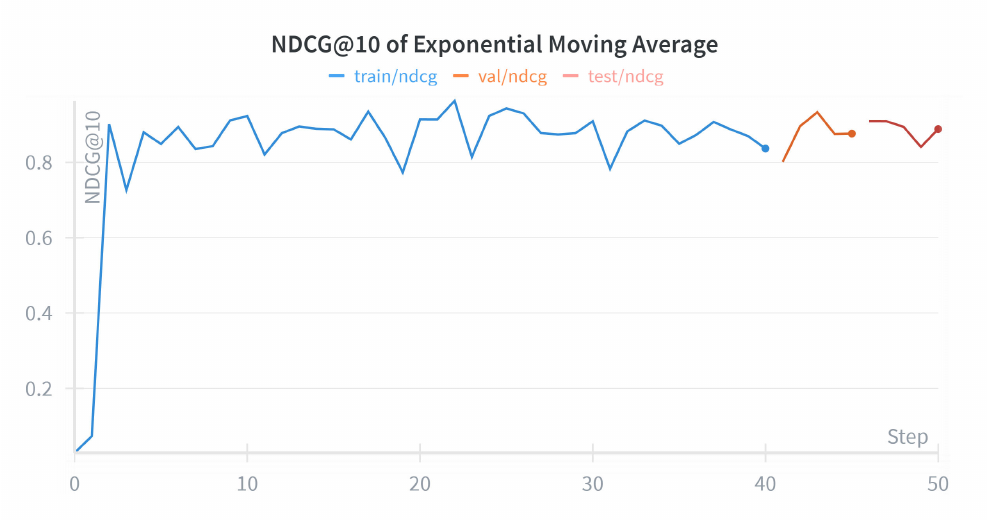}
	\caption{NDCG@10 over time for EMA}
	\label{exper:vis:ema}
\end{figure}

% Persistent Forecast
\begin{figure}[tbp]
	\centering
	\includegraphics[width=\linewidth]{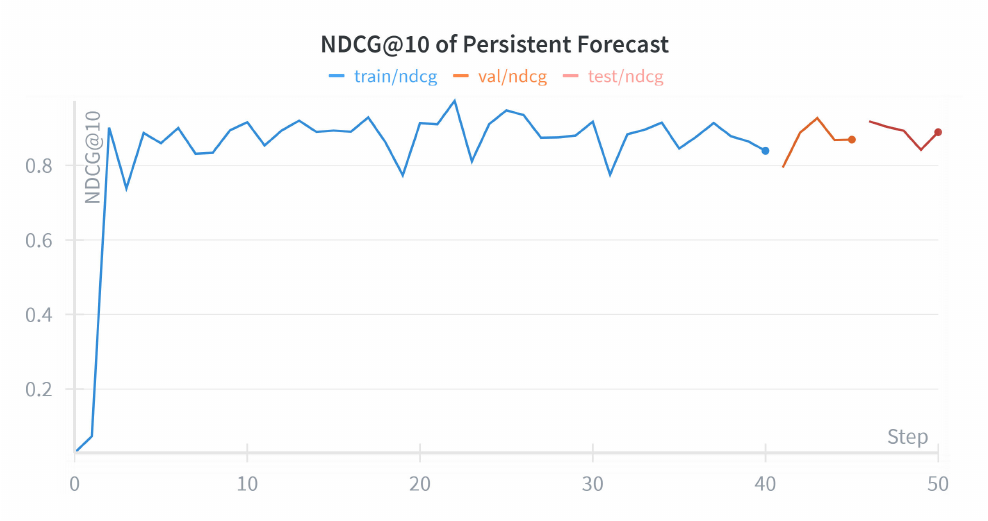}
	\caption{NDCG@10 over time for PF}
	\label{exper:vis:pf}
\end{figure}

% Moving Average
\begin{figure}[tbp]
	\centering
	\includegraphics[width=\linewidth]{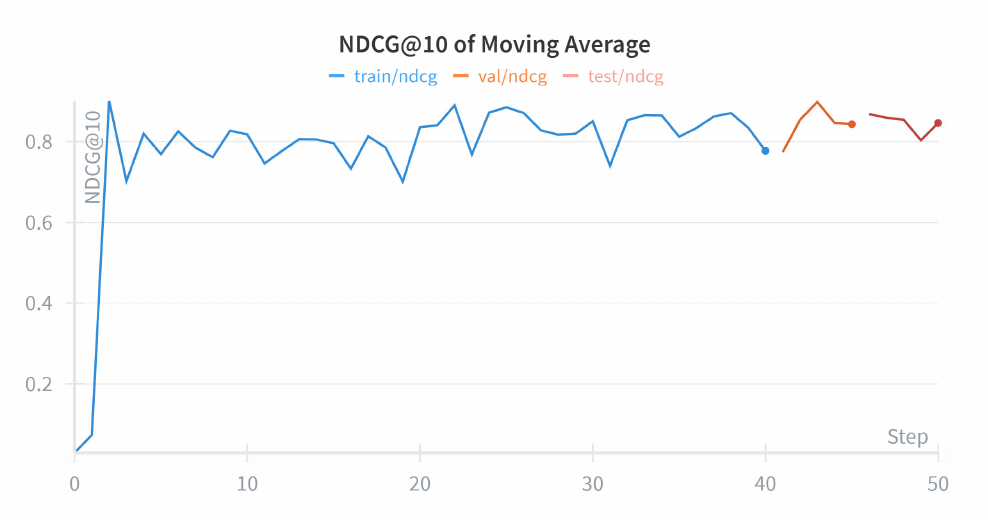}
	\caption{NDCG@10 over time for MA}
	\label{exper:vis:ma}
\end{figure}

% TGNv2_with_features
\begin{figure}[tbp]
	\centering
	\includegraphics[width=\linewidth]{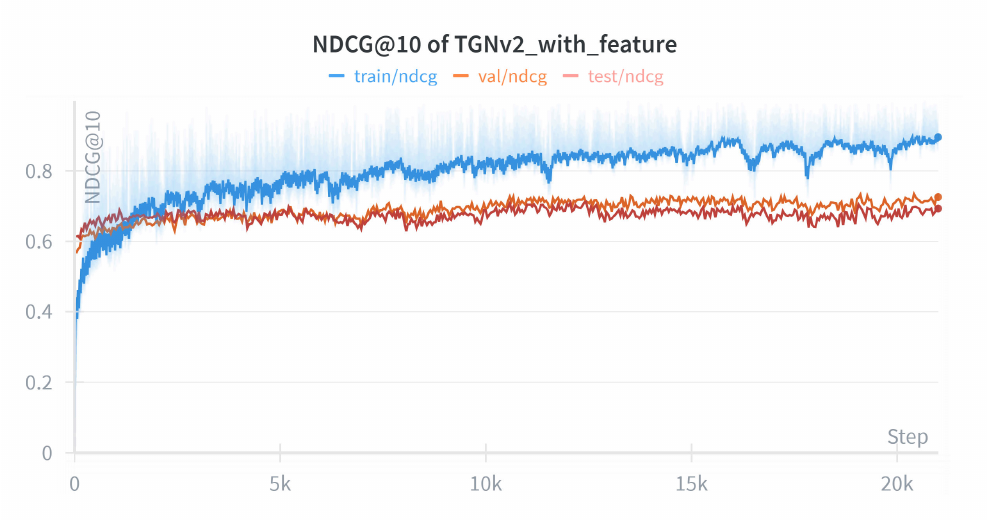}
	\caption{NDCG@10 over time for TGNv2 with node features}
	\label{exper:vis:tgnv2feat}
\end{figure}

% TGN_with_features
\begin{figure}[tbp]
	\centering
	\includegraphics[width=\linewidth]{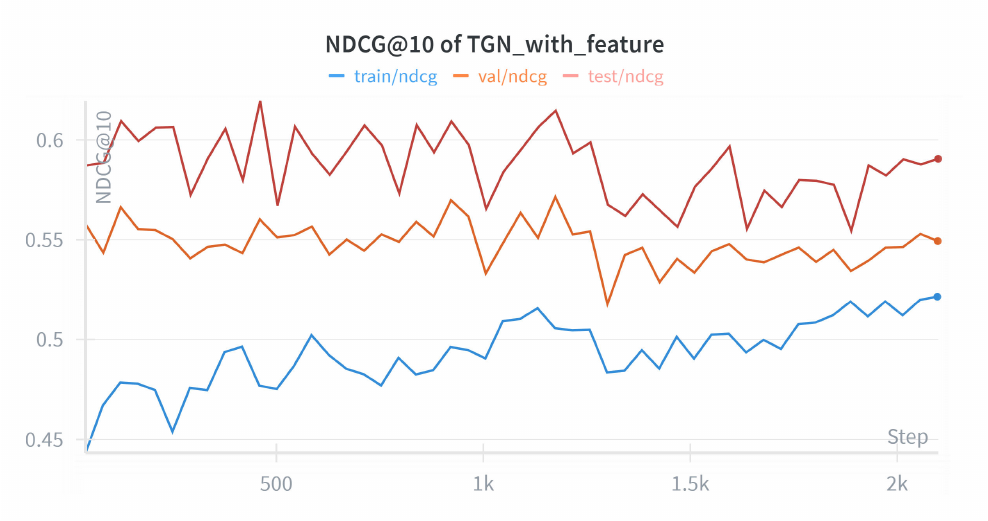}
	\caption{NDCG@10 over time for TGN with node features}
	\label{exper:vis:tgnfeat}
\end{figure}

% DyGFormer_with_features
\begin{figure}[tbp]
	\centering
	\includegraphics[width=\linewidth]{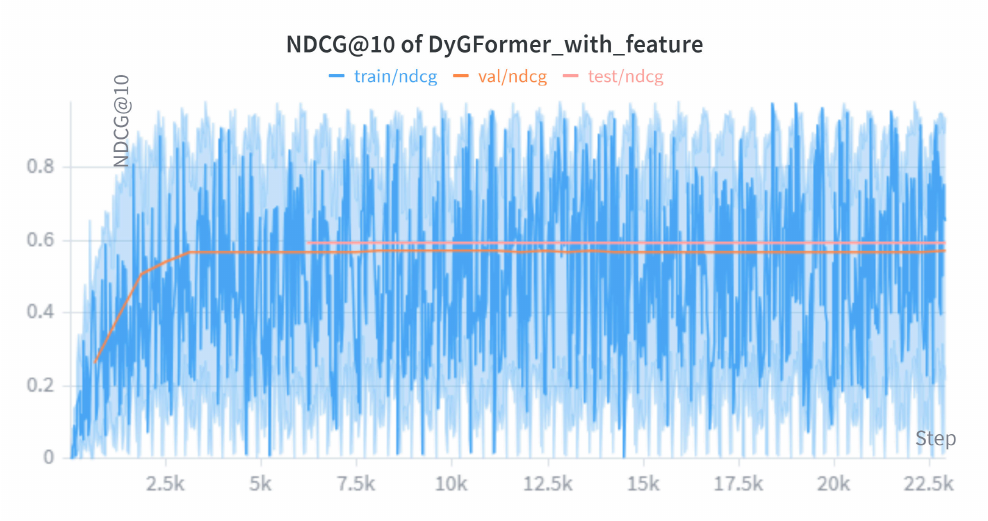}
	\caption{NDCG@10 over time for DyGFormer with node features}
	\label{exper:vis:dygformerfeat}
\end{figure}

% DyRep_with_features
\begin{figure}[tbp]
	\centering
	\includegraphics[width=\linewidth]{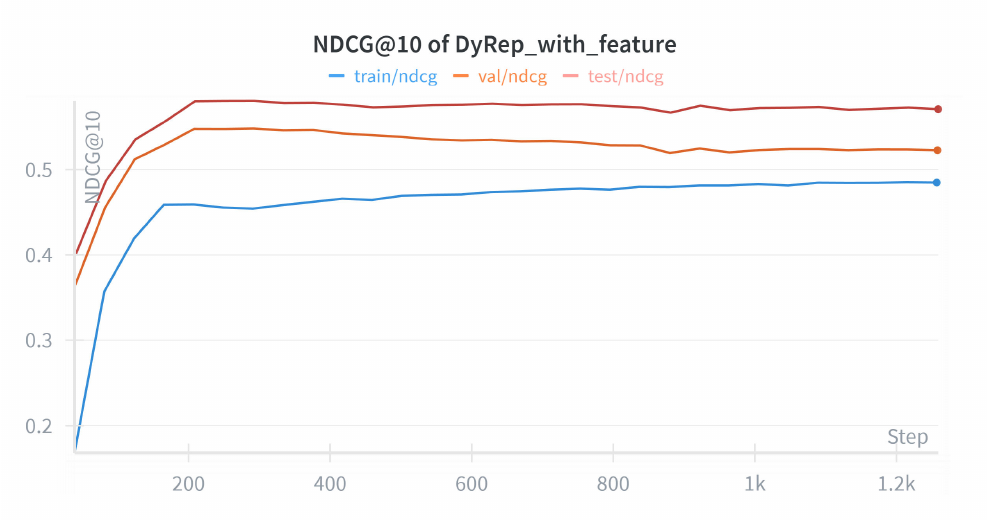}
	\caption{NDCG@10 over time for DyRep with node features}
	\label{exper:vis:dyrepfeat}
\end{figure}

\section{Conclusion}
\label{conc}
This work introduces a novel application of temporal graph machine learning to the prediction of institutional equity holdings from SEC Form 13F filings. By framing the problem as forecasting node affinities (percentage portfolio allocations) on a discrete-time temporal bipartite graph of managers and securities, we establish a rigorous, reproducible benchmark for this important financial prediction task. While our experiments focus on institutional equity holdings from SEC Form 13F filings, the underlying proposed node affinity prediction formulation is more general and can be applied to other regulatory filings involving a broader range of financial instruments and asset classes involving both equity and non-equity.

Our comprehensive evaluation on a random sample of 99 institutional investment managers and the full S\&P 500 index (503 securities, 209,351 temporal edges) reveals several key insights. First, NAVIS emerges as the clear state-of-the-art, achieving a test NDCG@10 of 0.9127 with features and 0.9121 without, dominating all dynamic graph representation learning models by a substantial margin, and all heuristic methods. Second, a simple Exponential Moving Average delivers remarkably strong performance (test NDCG@10 = 0.8882), outperforming all sophisticated dynamic graph representation learning models except NAVIS, and all other heuristic methods except Persistent Forecast, which achieves test NDCG@10 = 0.8891. This underscores the high persistence and smoothness inherent in institutional portfolio changes, where recent allocations are highly predictive of future ones. The top-performing affinity predictions---particularly those from NAVIS and the heuristic methods---nevertheless capture persistent and economically intuitive allocation patterns, with many managers exhibiting stable preferences toward particular subsets of securities that evolve gradually over time and remain informative about future portfolio adjustments.

Third, while domain-specific node features provide consistent gains across learned models, the improvement is slight (typically <1.2\%). This suggests that the temporal and structural signals in the 13F ownership graph already capture most of the predictive information for institutional holdings forecasting, allowing advanced models such as NAVIS to perform exceptionally well even without node features. The fact that NAVIS performs exceptionally well even in the featureless setting highlights the power of temporal graph representation learning for financial time-series prediction. Nevertheless, these findings are informative, as they suggest that further performance improvements are more likely to come from developing more informative financial features or more expressive temporal graph representation learning architectures than from simply increasing the number of handcrafted node attributes.

These findings have important implications. For practitioners, the strong performance of simple statistical methods like EMA indicates that sophisticated dynamic graph representation learning models may offer only marginal benefits in highly persistent domains like institutional holdings—a reminder that baselines should never be underestimated. However, this observation should not be interpreted as diminishing the value of temporal graph representation learning. Financial markets often exhibit structural changes, regime shifts, abrupt market events, and varying degrees of persistence across markets, datasets, asset classes, and regulatory disclosure frameworks. In such settings, learned temporal graph representations are well suited to capturing complex structural and temporal dependencies that may not be adequately modeled by simple persistence-based heuristics, making them a promising direction for more challenging and less persistent prediction problems. For researchers, our benchmark demonstrates the value of bringing state-of-the-art temporal graph representation learning models to finance, while revealing opportunities to develop more informative financial features and more expressive model architectures.

By framing institutional holdings prediction as a node affinity prediction problem and evaluating a suite of temporal graph models on the TGB node property prediction track---both with and without node features---this work establishes a rigorous, reproducible benchmark for applying temporal graph machine learning to institutional holdings dynamics, while providing a foundation for future research on ownership concentration, portfolio evolution, institutional herding, crowding risks, price pressure, and latent institutional demand in financial markets.

\section{Limitations}
The proposed benchmark is constructed from a randomly sampled subset of institutional investment managers rather than the complete universe of SEC Form 13F filers. Specifically, 99 managers were selected due to computational resource constraints, particularly the lack of access to GPU hardware required for training and evaluating multiple temporal graph representation learning models under a unified benchmarking framework. While this sampling reduces the size of the benchmark, it preserves the realistic temporal characteristics of institutional portfolio evolution and enables reproducible comparison across a diverse set of methods. In contrast, the benchmark includes all 503 constituents of the S\&P 500 index, providing complete coverage of one of the most widely followed segments of the U.S. equity market.

The benchmark is further limited to publicly available quarterly Form 13F disclosures, which primarily report long equity positions and therefore do not capture short positions, derivatives, or many non-equity asset classes. Moreover, the quarterly reporting frequency inherently limits the temporal resolution of the benchmark.

Despite these limitations, the proposed methodology is not restricted to the specific benchmark presented in this work. Both the heuristic baselines and, in particular, the dynamic graph representation learning models are inherently scalable and can be applied to substantially larger datasets as computational resources permit. Furthermore, the proposed node affinity prediction formulation is independent of SEC Form 13F and can naturally be extended to other regulatory disclosure datasets, such as Form 13D, Form 13G, and Form N-PORT, as well as to portfolios containing multiple asset classes. Unlike heuristic methods that primarily exploit persistence in historical allocations, dynamic graph representation learning models learn evolving structural and temporal dependencies directly from the underlying graph. Consequently, they are expected to remain applicable in settings exhibiting greater market volatility, structural changes, or weaker portfolio persistence, where richer temporal dynamics become increasingly important.

\section{Future Work}
\label{dis}
Future work could evaluate the proposed framework on larger datasets and additional regulatory filings, including Form 13D, Form 13G, and Form N-PORT, as well as portfolios containing multiple asset classes. Such settings are likely to exhibit more complex temporal dynamics and varying degrees of persistence, providing a broader evaluation of temporal graph representation learning models beyond highly persistent institutional holdings. Another direction is to develop more powerful architectures and richer financial features, incorporate short positions where available, or explore multi-step forecasting. Furthermore, generalizing the node affinity prediction task from dynamic graphs to dynamic hypergraphs represents another promising research direction for temporal graph machine learning and financial applications, as many financial interactions involve higher-order relationships that cannot be adequately captured by pairwise graphs. The reproducible setup and public TGB integration we provide lay a solid foundation for such advances and for future research on ownership concentration, portfolio evolution, institutional herding, crowding risks, price pressure, and latent institutional demand in financial markets.

\bibliographystyle{elsarticle-harv}

\end{document}